# A Single-Target License Plate Detection with Attention

Wenyun Li and Chi-Man Pun[*]

*Abstract—* With the development of deep learning, Neural Network is commonly adopted to the License Plate Detection (LPD) task and achieves much better performance and precision, especially CNN-based networks can achieve state of the art RetinaNet[1]. For a single object detection task such as LPD, modified general object detection would be time-consuming, unable to cope with complex scenarios and a cumbersome weights file that is too hard to deploy on the embedded device.

## I. Introduction and Proposed Method

In modern intelligent transportation system such as traffic control, vehicle surveillance and car park management system, license plate number will link to a large amount of information such as ownership, vehicle status and driver's record [1]. So license plate detection and recognition(LPDR) is a very essential task, and license plate detection plays a more important role. The performance of detection module depends on the high accuracy and precision of image classification and recognition [2-4]. In nowadays, popular LPDR[6] mainly includes end to end and modeled to detection and recognition task. When compared to end to end implementation, the separated detection and recognition are more common and widely suitable [9]. So we also implement divided design and more focus on detection (LPD).

In this paper, we proposed a network based on RetinaFace[5] network, which is able to switch to different backbone like ResNet-50[7] and MobileNet[8], so it could easily trade in accuracy and speed of detection and suitable for embedded device. Besides, we also added the spatial and channel attention to the model and improved IoU loss. The main contributions of this work are as follows:

Based on the good performance single target detection network, to avoid the general object detection networks shortage, we use an improved Retina network to make the prediction of the bbox and the four corner points of the license plate for future recognition.

We add a tiny but useful attention module to the network and make the prediction network more sensitive to the license plate and its spatial information.

We adapt the CIoU loss in bbox IoU loss in detection, and finally our model can achieve 99.6% accuracy in the CCPD[6] dataset.

[*]Corresponding author

This work was partly supported by the University of Macau (File no. MYRG2018-00035-FST, MYRG2019-00086-FST) and the Science and Technology Development Fund, Macau SAR (File no. 0034/2019/AMJ, 0087/2020/A2).

Wenyun Li and Chi-Man Pun are with the Department of Computer and Information Science, University of Macau, Macau, China (e-mail: cmpun@umac.mo).

## II. Results and Discussion

For IoU choice of our design, unlike the design of RetinaNet using Smooth L1 Loss as bbox loss, which represents the location of four coordinates (box's center and its width and height). When bbox and target box has no union part, or IoU= 0, lower loss will equal to the higher loss, the optimization will not work properly. We choose CIoU as our bbox loss.

On the other hand, the license plate is a standard flat significant object, so by adding an attention mechanism would make the model more sensitive to the license plate, here we refer to the design of, employ the spatial attention and channel attention simultaneously for improving the performance of ResNet features.

| Accuracy (%) | CCPD-Base | CCPD-weather | CCPD-tilt | CCPD-Rotate | CCPD-fn | CCPD-db | CCPD-Challenge |
|---|---|---|---|---|---|---|---|
| VGG16[5] | 95.3 | 94.1 | 93.8 | 92.0 | 91.9 | 93.5 | 93.3 |
| RPnet[2] | 96.8 | 96.1 | 95.0 | 92.8 | 93.6 | 95.7 | 94.6 |
| CMnet[6] | 98.8 | 98.6 | 98.5 | 97.9 | 96.5 | 97.3 | 96.8 |
| Our | 99.8 | 99.8 | 99.6 | 99.7 | 97.3 | 97.9 | 99.2 |

Table 1. Modified Retina LPD system experiment results.

As shown in Table 1, our proposed model can achieve more accurate performance in the CCPD dataset.